%% file: paper.tex
\documentclass{MIPRO}

\usepackage[nocompress]{cite}
\usepackage{amsmath,amssymb,amsfonts}
\usepackage{algorithmic}
\usepackage{graphicx}
\usepackage{textcomp}
\usepackage{xcolor}
\usepackage[T1]{fontenc}
\usepackage[]{flushend}
\usepackage{multirow}
\usepackage{multicol}
\usepackage{array}
\usepackage{url}
\usepackage{hyperref}
\usepackage{graphicx}
\graphicspath{ {./figures/} }

\begin{document}

\title{Towards AI Evaluation in Domain-Specific RAG Systems: The AgriHubi Case Study}

\author{
\IEEEauthorblockN{
Md.~Toufique Hasan\IEEEauthorrefmark{1},
Ayman Asad Khan\IEEEauthorrefmark{1},
Mika Saari\IEEEauthorrefmark{1},
Vaishnavi Bankhele\IEEEauthorrefmark{1},
Pekka Abrahamsson\IEEEauthorrefmark{1}
}
\IEEEauthorblockA{
\IEEEauthorrefmark{1}Faculty of Information Technology and Communication Sciences, Tampere University\\
Tampere, Finland\\
\{mdtoufique.hasan, ayman.khan, mika.saari, vaishnavi.bankhele, pekka.abrahamsson\}@tuni.fi
}
}

\maketitle

\begin{abstract}
Large language models show promise for knowledge-intensive domains, yet their use in agriculture is constrained by weak grounding, English-centric training data, and limited real-world evaluation. These issues are amplified for low-resource languages, where high-quality domain documentation exists but remains difficult to access through general-purpose models. This paper presents \textit{AgriHubi}, a domain-adapted retrieval-augmented generation (RAG) system for Finnish-language agricultural decision support. AgriHubi integrates Finnish agricultural documents with open PORO family models and combines explicit source grounding with user feedback to support iterative refinement. Developed over eight iterations and evaluated through two user studies, the system shows clear gains in answer completeness, linguistic accuracy, and perceived reliability. The results also reveal practical trade-offs between response quality and latency when deploying larger models. This study provides empirical guidance for designing and evaluating domain-specific RAG systems in low-resource language settings.
\end{abstract}

\renewcommand\IEEEkeywordsname{Keywords}

\section{Introduction}
\label{sec:introduction}
\input{01_introduction}

\section{Background}
\label{sec:background}
\input{02_background}

\section{System Design and Architecture}
\label{sec:system_design}
\input{03_system_design_and_architecture}

\section{Implementation and Iterations}
\label{sec:implementation}
\input{04_implementation_and_iterations}

\section{Evaluation and Results}
\label{sec:evaluation}
\input{05_evaluation_and_results}

\section{Discussion}
\label{sec:discussion}
\input{06_discussion}

\section{Future Work}
\label{sec:future}
\input{07_future_work}

\section{Summary}
\label{sec:summary}
\input{08_summary}

\section*{Acknowledgments}
This work was funded by the European Regional Development Fund and the Regional Council of Satakunta. 

\bibliographystyle{IEEEtran}
\bibliography{references}

\end{document}

%% file: 01_introduction.tex
Agricultural decision making increasingly relies on large and fragmented sources of information, including research reports, regulatory documents, advisory guidelines, and environmental records. Although much of this material is publicly available, practitioners often struggle to access, interpret, and apply it effectively in practice. At the same time, recent advances in generative artificial intelligence (GenAI) have created new opportunities for retrieving and presenting domain knowledge through conversational interfaces powered by large language models (LLMs) \cite{nguyen2025generative}.

General-purpose LLMs are poorly suited for specialized domains. They often produce confident but weakly grounded answers because they fail to connect domain facts consistently \cite{wang2024reasoning}. In agriculture, where advice depends on complex climate–soil relationships, this results in frequent hallucinations due to limited domain-specific training data.

Retrieval-augmented generation (RAG) has emerged as a promising approach to address these shortcomings by combining document retrieval with generative models. Prior work has shown that grounding responses in curated agricultural sources improves both relevance and reliability. Samuel et al. \cite{samuel2025agrollm} demonstrated that retrieving context from specialized agricultural literature yields more accurate and useful answers than standalone generation. Fanuel et al. \cite{fanuel2025agriregion} further showed that prioritizing region-specific documents during retrieval significantly improves trustworthiness, underscoring the importance of localized knowledge for agricultural applications.

However, a critical gap remains. Most existing agricultural RAG systems focus on English-language data, and there is limited empirical evidence on domain-specific RAG systems designed for low-resource languages and evaluated in real-world settings. Finnish-language agricultural decision support, in particular, has received little attention, despite the availability of high-quality national documentation and the emergence of Finnish-capable open LLMs such as the PORO family.

This paper addresses this gap by presenting \textit{AgriHubi}, a domain-adapted RAG system that integrates Finnish agricultural documents with open LLMs from the PORO family. AgriHubi is designed for real-world use, with explicit source grounding and a built-in feedback mechanism that supports iterative refinement. The work is aligned with national AI and sustainability initiatives in Finland, including \textit{Hiilestä kiinni} and \textit{Ruokavirasto}, which emphasize responsible data use and practical decision support.

The study is guided by the following research questions:
\begin{itemize}
    \item \textbf{RQ1:} How can RAG architecture be adapted to effectively support Finnish-language agricultural data using open-source LLMs?
    \item \textbf{RQ2:} How does model choice within the PORO family affect response quality, latency, and linguistic accuracy in the Finnish agricultural context?
    \item \textbf{RQ3:} How does iterative refinement based on user feedback influence the reliability and usability of a deployed RAG system over time?
\end{itemize}

To answer these questions, we have designed, implemented, and evaluated AgriHubi through multiple development iterations and two structured user evaluation rounds. 

%% file: 02_background.tex
This section examines the prior work on multilingual language models, retrieval-augmented generation, and evaluation practices in agricultural AI, highlighting persistent gaps in low-resource language support and empirically validated domain-specific deployments.

The dominance of English in foundational model development has created a prominent performance disparity for smaller languages. To address this English bias, Lai et al. \cite{lai2024llms} introduced xLLMs-100, demonstrating that scaling multilingual instruction tuning across 100 languages substantially enhances cross-lingual alignment. However, for languages with limited training corpora, language-specific optimization remains vital. Luukkonen et al. \cite{luukkonen2025poro} showed that multilingual training can benefit low-resource languages, with their Poro 34B model outperforming previous Finnish-focused models despite data scarcity. This aligns with the survey by Ali and Pyysalo \cite{ali2024survey}, which suggests that, while multilingual models are improving, language-specific models typically yield superior performance when sufficient high-quality data is available.

Beyond linguistic challenges, general-purpose models often struggle with the specialized knowledge required for agricultural decision support. A comprehensive review by Yin et al. \cite{yin2025foundation} highlights that adapting foundation models to specific farming contexts, particularly through retrieval-augmented and multimodal approaches, is essential for tasks such as crop disease detection and yield prediction. Data quality plays a pivotal role in this adaptation; Jiang et al. \cite{jiang2025knowledge} found that constructing knowledge bases from professional literature yields significantly better performance than relying on general internet data. However, localized deployment faces hurdles and, as Owiti and Kipkebut \cite{owiti2025enhancing} noted, the lack of high-quality annotated datasets in native languages remains a primary bottleneck for training models that genuinely understand local farming terminology.

To mitigate these limitations, recent research has moved toward more sophisticated architectures to enhance reliability. Zhang et al. \cite{zhang2025beefbot} demonstrated with BeefBot that combining fine-tuning with retrieval-augmented generation (RAG) significantly reduces hallucinations compared to general-purpose models. Further architectural innovations include the ``Tri-RAG'' framework proposed by Yang et al. \cite{yang2025agrigpt}, which integrates dense retrieval, sparse retrieval, and knowledge graphs to improve reasoning on complex tasks. Moving toward agentic workflows, Yang et al. \cite{yang2024shizishangpt} developed ShizishanGPT, confirming that integrating external tools for specialized tasks, such as phenotype prediction, considerably outperforms standard model capabilities.

Despite these architectural advancements, gaps remain in evaluation and practical deployment. Brown et al. \cite{brown2025systematic} note that evaluation practices are inconsistent, often relying on basic metrics such as Exact Match rather than assessing retrieval quality or grounding effectiveness. Furthermore, Hasan et al. \cite{hasan2026engineering} argue that, while theoretical RAG research is abundant, there is a lack of empirical studies on domain-specific systems deployed in real-world settings. Addressing practical deployment challenges, Suvitie et al. \cite{Suvitie2026MIPRO} showed that hybrid human-in-the-loop workflows can greatly reduce data creation costs. Their approach achieved a cost of about \$0.25 per 100 pages. This makes it possible to build high-quality domain datasets without large annotation teams.

These limitations motivated the present work, which focuses on the design, deployment, and evaluation of a domain-specific, Finnish-language RAG system for real-world agricultural decision support.

%% file: 03_system_design_and_architecture.tex
This section describes the design of AgriHubi and how its components work together. The system is organized around data ingestion, retrieval, model inference, and user interaction, forming a complete RAG pipeline that transforms agricultural documents into context-aware answers. The following subsections outline the overall architecture, key components, and underlying technologies.

\subsection{Overall System Architecture}

AgriHubi follows a RAG pipeline with four core parts: document store, retriever, generative model, and user interface. When a user submits a query, the system embeds it and searches a FAISS index built from preprocessed agricultural PDFs. The highest-scoring text chunks are grouped into a context window and sent along with the query to the selected model (\texttt{Llama\,3.2}, \texttt{PORO-34B}, \texttt{PORO-2-8B}, or \texttt{PORO-2-70B}) through the \texttt{Farmi}/\texttt{GPT-Lab} APIs.

\begin{figure}[ht]
    \centering
    \includegraphics[width=\linewidth]{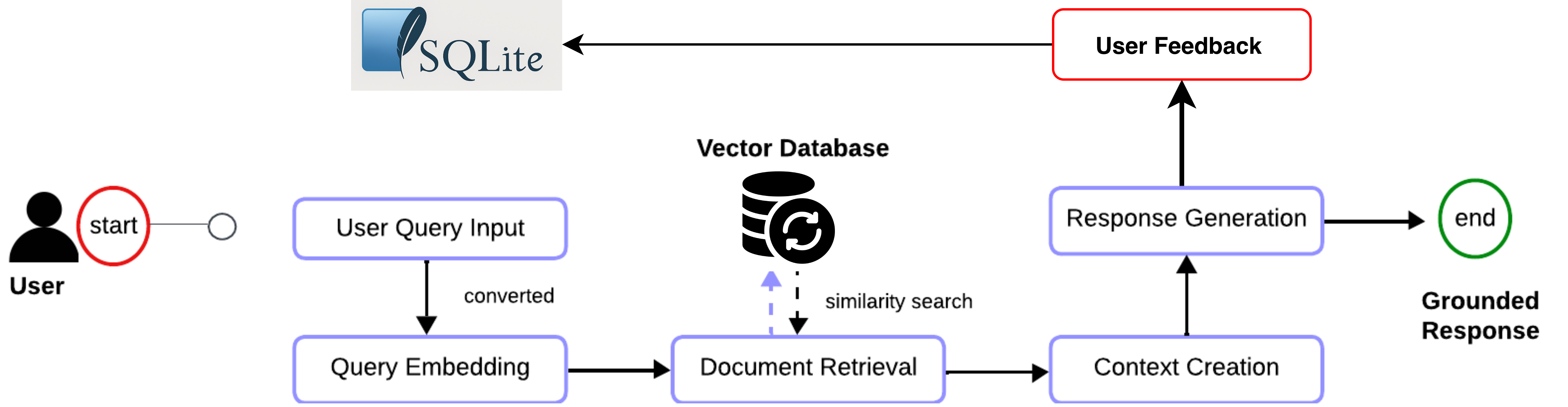}
    \caption{AgriHubi RAG workflow from query embedding to retrieval, context construction, generation, and feedback storage.}
    \label{fig:rag_workflow}
\end{figure}

The Streamlit interface streams the model’s response back to the user and logs the query, retrieved passages, and ratings in a local SQLite database. Figure~\ref{fig:rag_workflow} shows the full workflow used in deployment.

\subsection{Components}

\begin{figure*}[t]
    \centering
    \includegraphics[width=\textwidth]{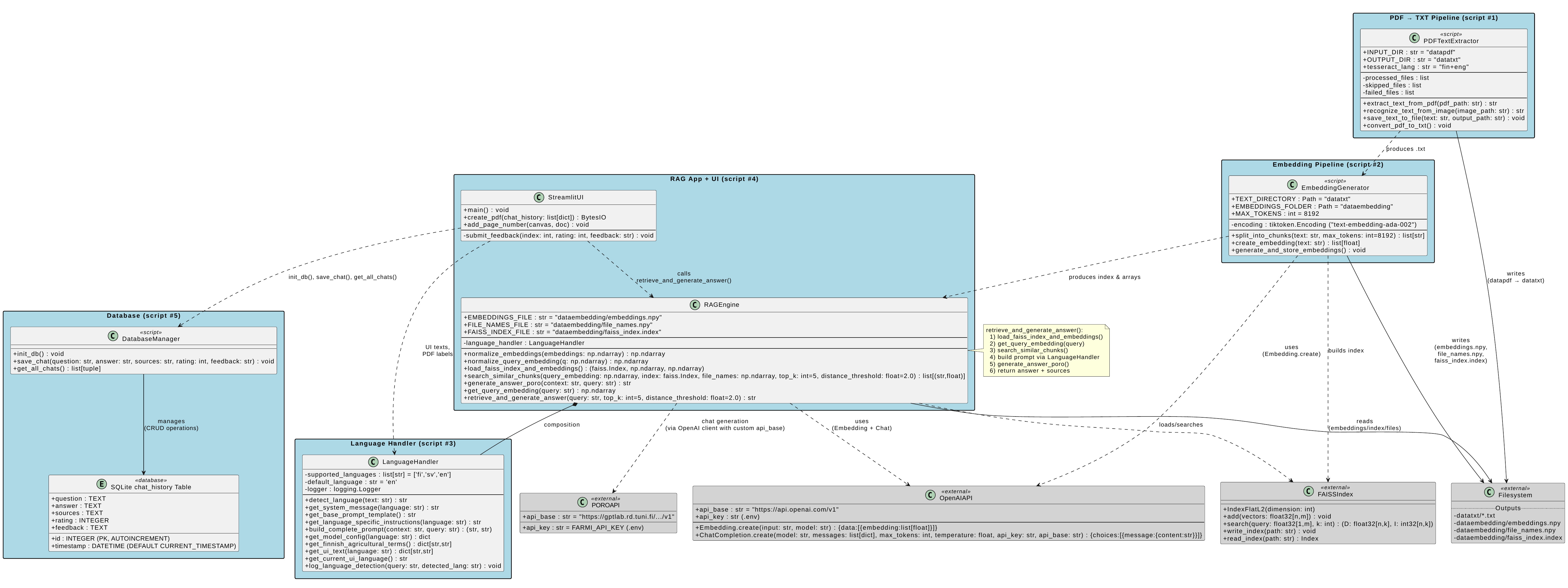}
    \caption{UML Class Diagram showing the main modules, data flows, and interactions within the AgriHubi system.}
    \label{fig:uml_class_diagram}
\end{figure*}

AgriHubi is built from modular components that handle ingestion, retrieval, generation, and user interaction. The UML Class Diagram in Figure~\ref{fig:uml_class_diagram} shows how these pieces fit together.

\subsubsection{Data Ingestion and Preprocessing}

This component converts raw agricultural PDF documents into structured, searchable text. The pipeline includes:
\begin{itemize}
\raggedright
    \item PDF text extraction with OCR fallback for scanned documents,
    \item segmentation of extracted text into coherent chunks,
    \item metadata tagging to preserve document identity and support traceability.
\end{itemize}
The preprocessing scripts \path{pdf_to_txt.py} and \path{txt_to_embedding.py} implement these steps and are part of the internal AgriHubi codebase.

\medskip
\subsubsection{Embedding and Indexing}

The system embeds text chunks using OpenAI’s \texttt{text-embedding-ada-002} model and stores them in a FAISS index. The setup uses L2-normalized vectors, cosine similarity scoring, and then persistent files in \texttt{/dataembedding/}. This enables fast and reliable similarity search.

\medskip
\subsubsection{Generative Model Integration}

AgriHubi supports multiple Finnish-capable LLMs across its development timeline:
\begin{itemize}
\raggedright
    \item \texttt{Llama\,3.2} for early testing,
    \item \texttt{PORO-34B} for better Finnish output,
    \item \texttt{PORO-2-8B} for multilingual robustness,
    \item \texttt{PORO-2-70B} for the strongest final performance.
\end{itemize}

The scripts \path{main_poro34b.py}, \path{main_poro2_8b.py}, and \path{main_poro2_70b.py} combine retrieved context with prompt templates and call the models through the OpenAI/Farmi APIs.

\medskip
\subsubsection{User Interface}

The Streamlit-based user interface supports Finnish, Swedish, and English, with automatic translation of interface text handled by a centralized language handler. User interactions follow a chat-style format with streamed responses from the model. The interface also allows users to export conversation histories as PDF files and provides a built-in five-point rating mechanism for collecting structured feedback on answer quality.

\medskip
\subsubsection{Database and Feedback Loop}

The SQLite database (\texttt{chat\_history.db}) stores each query, the retrieved document chunks, the model’s response, the user rating and any written feedback, along with timestamps and model identifiers. This record of interactions formed the core of the feedback loop that guided improvements throughout development.

\subsection{Technical Stack}

AgriHubi uses a Python-based architecture built for clarity, modularity, and reproducibility.

\medskip
\subsubsection{Core Technologies}

\begin{itemize}
\raggedright
    \item \texttt{Python\,3.11.11} for the main application,
    \item FAISS for vector retrieval,
    \item OpenAI embedding for text representation,
    \item Streamlit for the user interface,
    \item SQLite for persistent evaluation data,
    \item \texttt{dotenv} for secure API key handling.
\end{itemize}

\subsubsection{Model Hosting}

The PORO models (\texttt{PORO-2-70B}, \texttt{PORO-2-8B}, \texttt{PORO-34B}) run on \texttt{GPT-Lab}/\texttt{Farmi} servers, so no local GPUs are required.

\subsubsection{Deployment, and Supporting Utilities}

AgriHubi can be deployed locally or on laboratory servers using the dependencies listed in \texttt{requirements.txt}. Its lightweight, modular design supports containerized setups where frontend and backend components run independently. The system also includes utilities for text preprocessing, database management, and prompt and language handling, with architectural diagrams maintained under \texttt{/diagram/}. These components support maintainability, iterative development, and future system extensions.

%% file: 04_implementation_and_iterations.tex
AgriHubi was developed through eight iterations between January and August~2025. Each iteration introduced targeted improvements based on testing and user feedback. Over time, the system evolved from an initial RAG prototype into a stable, Finnish-optimized platform with improved context handling, terminology, and retrieval reliability.

\subsection{Iterations 1--3: Establishing the Core System}

The first three iterations focused on building the core RAG pipeline. We added PDF extraction, OCR for scanned documents, text chunking, embedding generation, and a FAISS index for semantic retrieval. The early system relied on \texttt{Llama\,3.2}. It worked for initial testing but struggled with Finnish agricultural terminology. Iteration~2 introduced \texttt{PORO-34B}, which immediately improved fluency and domain coverage. We also released the first Streamlit interface so internal testers could try the system directly. In Iteration~3, the document collection was expanded and Finnish terminology refined based on tester feedback. These stages form the foundation of the repository’s \texttt{main} branch.

\subsection{Iterations 4--6: Feedback and Model Refinement}

Iteration~4 added a built-in feedback mechanism with a five-point rating scale stored in SQLite. This gave us clear insight into answer quality. The collected data showed issues with accuracy, incomplete reasoning, and occasional ambiguity. Iteration~5 included the first structured external evaluation (67 responses). Almost half of the answers were rated 1--2, which led to changes in the retrieval strategy, chunking logic, and similarity thresholds. During this phase, we also tested the \texttt{Viking-33B} model, but its responses were short and unstable, so it was removed. Iteration~6 transitioned the system to \texttt{PORO-2-8B}, which produced more stable Finnish and Swedish output and handled terminology more consistently. Preprocessing and embedding generation were also improved. These updates are captured in the \texttt{v6.0} branch.

\subsection{Iterations 7--8: Finalizing the System}

Iteration~7 introduced \texttt{PORO-2-70B}, the strongest model in terms of accuracy and completeness. Its answers were longer, clearer, and more grounded in retrieved context. Supporting it required backend upgrades, including larger model calls, improved streaming, and updated timeout logic. These appear in the \texttt{v7.0} branch in the internal AgriHubi codebase. Iteration~8 concentrated on usability and system stability. We replaced filename-based retrieval with full-chunk retrieval, increased the maximum answer length from 700 to 2000 tokens, improved error handling, and fully localized the interface into Finnish. These updates addressed repeated user requests for more complete answers. The \texttt{v8.0} changes led to a clear jump in evaluation scores in the second testing round (47 responses), especially regarding completeness and clarity.

User feedback guided most design decisions. Testers reported problems with terminology, missing context, unclear answers, and slow responses. Their ratings led to changes in retrieval settings, model selection, interface design, and server setup. As a result, the system was improved through repeated testing and refinement rather than a fixed development process.

%% file: 05_evaluation_and_results.tex
This section explains how we evaluated AgriHubi and what the results show. Two user studies and several internal tests reveal how the system’s accuracy, fluency, and responsiveness changed over time. Together, these findings highlight where the system has been improved and where limitations remain.

\subsection{Methodology}

AgriHubi was evaluated in two rounds in April and August 2025. In both rounds, users asked real agricultural questions and rated the answers on a five-point Likert scale (1 = poor, 5 = excellent). The system also logged each query, retrieved document segments, model responses, and timestamps in a local SQLite database, enabling consistent comparison across iterations. The April evaluation tested the \texttt{PORO-34B} version and collected 67 ratings. The August evaluation assessed the updated system using \texttt{PORO-2-70B}, and in some cases \texttt{PORO-2-8B}, with 47 ratings. Question topics were kept similar to support direct comparison between the two rounds.

Across both phases, four primary evaluation metrics were used:
\begin{itemize}
    \item user-perceived quality via Likert ratings,
    \item end-to-end response latency,
    \item factual accuracy and completeness of generated answers,
    \item fluency of Finnish agricultural terminology.
\end{itemize}

These were complemented by internal system measurements and qualitative feedback from users.

\subsection{Quantitative Results}

\textbf{Round 1 (April 2025, \texttt{PORO-34B}).} Table~\ref{table:likert_comparison} summarizes the 67 collected ratings. The distribution was as follows: 15 responses (22\%) with rating~1, 16 responses (24\%) with rating~2, 17 responses (25\%) with rating~3, 17 responses (25\%) with rating~4, and 2 responses (3\%) with rating~5. Almost half of the answers (46\%) therefore fell into the lowest two categories, while only 3\% achieved the top score. Testers most often reported missing details, weak factual grounding, and inconsistent Finnish terminology.

\textbf{Round 2 (August 2025, \texttt{PORO-2-70B}).} Table~\ref{table:likert_comparison} also reports the 47 ratings collected with the updated system. Here, 9 answers (19\%) received rating~1, 9 answers (19\%) rating~2, 15 answers (32\%) rating~3, 4 answers (9\%) rating~4, and 10 answers (21\%) rating~5. Compared to Round~1, the share of low ratings (1--2) dropped from 46\% to 38\%, while the proportion of top-rated answers (score~5) increased from 3\% to 21\%. Ratings of 3 also increased from 25\% to 32\%, suggesting more stable and complete responses.

\begin{table}[ht]
\centering
\caption{Comparison of Likert Evaluation Results for April and August 2025}
\label{table:likert_comparison}
\begin{tabular}{|c|cc|cc|}
\hline
\multirow{2}{*}{\textbf{Rating}} 
& \multicolumn{2}{c|}{\textbf{April 2025}} 
& \multicolumn{2}{c|}{\textbf{August 2025}} \\ \cline{2-5}
& \textbf{Responses} & \textbf{\%} & \textbf{Responses} & \textbf{\%} \\ \hline
1 & 15 & 22\% & 9  & 19\% \\ \hline
2 & 16 & 24\% & 9  & 19\% \\ \hline
3 & 17 & 25\% & 15 & 32\% \\ \hline
4 & 17 & 25\% & 4  & 9\%  \\ \hline
5 & 2  & 3\%  & 10 & 21\% \\ \hline
\end{tabular}
\end{table}

\textbf{Comparison.} User ratings improved markedly between April and August. Low ratings (1--2) decreased from 46\% to 38\%, while top ratings (5) increased from 3\% to 21\%. Mid-range ratings (3) rose from 25\% to 32\%, and ratings of 4 declined from 25\% to 9\%.

\subsection{Internal Performance Evaluation}

Internal evaluations were conducted across all major model backends: \texttt{LLaMA}, \texttt{Viking-33B}, \texttt{PORO-34B}, \texttt{PORO-2-8B}, and \texttt{PORO-2-70B}. They were assessed along four criteria:
\begin{itemize}
    \item \textbf{User feedback:} average rating stability and distribution,
    \item \textbf{Qualitative errors:} factual accuracy and evidence grounding,
    \item \textbf{Latency:} end-to-end response time,
    \item \textbf{Domain coverage:} robustness of Finnish agricultural terminology.
\end{itemize}

\texttt{PORO-34B} delivered the first stable Finnish output but often produced incomplete answers. \texttt{PORO-2-8B} improved fluency with moderate latency, while \texttt{PORO-2-70B} achieved the highest accuracy at the cost of increased memory use and latency. The \texttt{Viking-33B} model was discontinued due to unstable, error-prone outputs, and \texttt{LLaMA} was used only as an early baseline. These internal results confirm a predictable trade-off: larger models yield higher answer quality but at the cost of increased latency and resource consumption.

\subsection{Qualitative Feedback}

Qualitative feedback helped explain the rating trends. In the first round, users often described answers as incomplete, overly general, inconsistent in Finnish terminology, or slow to generate. These comments match the low scores observed in April. In the second round, feedback shifted noticeably. Users described the system as clearer and more reliable, noting improved terminology, more natural Finnish, and better use of retrieved sources. Some issues remained, including occasional errors in highly technical questions and weaker performance for Swedish queries. Overall, the feedback shows a clear transition from a prototype to a usable decision-support tool, with steady gains in answer quality, completeness, and linguistic accuracy across iterations.

%% file: 06_discussion.tex
The evaluation shows that AgriHubi improved steadily across the development iterations. The strongest gains came from system-level refinements rather than model scaling alone. Improvements in retrieval quality, preprocessing, and context construction had a greater impact on answer quality than switching to larger models. This confirms that stable system design is critical for domain-specific RAG systems.

\subsection{Technical Observations}

Most technical improvements resulted from better retrieval and preprocessing. Early versions suffered from OCR noise, uneven chunking, and missing metadata, which prevented models from grounding answers correctly. Once these issues had been addressed, the PORO models used retrieved content more effectively and produced more complete responses.

A clear trade-off emerged between accuracy and latency. \texttt{PORO-2-70B} generated the most accurate and detailed answers but responded more slowly, especially under shared GPU load. Testers often associated slower responses with lower trust, showing that latency affects both usability and perceived reliability.

\subsection{Domain Impact}

AgriHubi showed clear progress in handling Finnish agricultural terminology. Testers reported more natural phrasing, more accurate use of domain concepts, and stronger alignment between answers and retrieved evidence. These gains resulted from both improved model capability and refinements in prompts and language handling.

Support for Swedish remained limited. Sparse training data and inconsistent document coverage led to unstable responses, reducing trust among bilingual users. This highlights the need for more balanced multilingual data and retrieval support in future versions.

\subsection{Limitations}

Several limitations influenced the evaluation. Hardware constraints limited the throughput of \texttt{PORO-2-70B}, and variable response times affected user ratings. The sample sizes of 67 and 47 responses provided useful insights but were insufficient for fine-grained statistical analysis. Privacy concerns also required care, as retrieved content sometimes included verbatim text from internal documents. In addition, differences in question types between evaluation rounds may have introduced minor scoring bias.

\subsection{Lessons Learned}

Three lessons stand out. First, incremental improvements driven by user feedback were more effective than large architectural changes. Second, human review was essential for identifying subtle errors in meaning, terminology, and context that automated signals could not reliably detect, especially in a low-resource language setting. Third, the lack of a consistent evaluation process across iterations made structured comparison difficult, pointing to the need for more systematic evaluation practices in future work.

Overall, the results show that AgriHubi evolved from an early prototype into a more reliable decision-support system. While gains in answer quality and user trust are clear, challenges related to latency, multilingual consistency, and production readiness remain open.

%% file: 07_future_work.tex
Future work will focus on expanding the system’s data ecosystem, strengthening contextual reasoning, and extending AgriHubi toward action-oriented decision support. A natural next step is deeper integration with external data sources, including sensor-based and API-driven inputs such as weather services, soil measurements, and environmental monitoring systems. Incorporating such real-time signals would allow the retrieval layer to reason not only over static documents but also over current conditions, enabling more situationally aware responses. Another important direction is the introduction of longer-term system memory. Persisting interaction history, retrieved context, and past recommendations would allow AgriHubi to maintain continuity across sessions and support cumulative reasoning over time. This form of contextual memory can improve consistency, reduce repeated explanations, and enable more informed follow-up guidance, especially in seasonal or longitudinal agricultural scenarios. Building on these capabilities, AgriHubi can gradually evolve from an information-centric RAG system into a broader decision-support platform. Future extensions could support lightweight planning tasks, scenario exploration, and recommendation workflows that connect retrieved knowledge with concrete actions, such as crop management suggestions or policy-compliant guidance. Achieving this vision will require careful data governance, controlled update cycles, and explicit handling of uncertainty when multiple data sources provide conflicting signals. Together, these directions will move AgriHubi beyond document question answering toward a grounded, context-aware agricultural AI system that supports real-world decision making.

%% file: 08_summary.tex
This paper introduced AgriHubi, a domain-specific retrieval-augmented generation system developed to support Finnish-language agricultural decision making. By integrating Finnish agricultural documents with open large language models from the PORO family, AgriHubi demonstrates how localized retrieval and language-aware generation can improve the reliability, clarity, and practical usefulness of AI-assisted decision support. The system was developed through eight iterative design cycles and evaluated in two structured user studies. Over time, AgriHubi progressed from an early prototype into a more stable and dependable system. Quantitative ratings and qualitative feedback show clear improvements in answer completeness, linguistic accuracy, and user trust, particularly after refinements to the retrieval pipeline and the transition to larger PORO models. The evaluation also revealed trade-offs between response quality and latency, reinforcing the central role of retrieval quality in grounding model outputs.

Beyond the Finnish agricultural context, this study offers practical insights into how domain-adapted RAG systems can be engineered, deployed, and evaluated in applied settings. The results highlight the importance of iterative refinement, user-centered evaluation, and transparent grounding when applying large language models in domain-specific and high-stakes environments. Taken together, the findings position AgriHubi both as a functional decision-support system and as a reference case for future research on multilingual, domain-specific retrieval-augmented generation.